\documentclass[letterpaper, 10 pt, conference]{ieeeconf}  % Comment this line out if you need a4paper

\IEEEoverridecommandlockouts                              % This command is only needed if 
\usepackage{amsmath} % assumes amsmath package installed
\usepackage{amssymb}  % assumes amsmath package installed
\usepackage{booktabs}
\usepackage{arydshln}
\usepackage[compress]{cite}
\usepackage{graphicx}
\usepackage{comment}
\usepackage{multirow}
\usepackage{amsfonts}
\usepackage{amssymb}
\usepackage[nolist]{acronym}

\usepackage{caption}

\title{\LARGE \bf
Dual-Control Frequency-Aware Diffusion Model for Depth-Dependent Optical Microrobot Microscopy Image Generation
}

\author{Lan Wei, Zongcai Tan, Kangyi Lu, Jian-Qing Zheng, Dandan Zhang
\thanks{Lan Wei, Zongcai Tan, Kangyi Lu, Dandan Zhang are with the Department of Bioengineering, Imperial-X AI Initiative, Imperial College London, London, United Kingdom. Jian-Qing Zheng is with CAMS-Oxford Institute, University of Oxford, Oxford, United Kingdom. Corresponding: d.zhang17@imperial.ac.uk.}
}
\begin{document}

\maketitle
\thispagestyle{empty}
\pagestyle{empty}

%%%%%%%%%%%%%%%%%%%%%%%%%%%%%%%%%%%%%%%%%%%%%%%%%%%%%%%%%%%%%%%%%%%%%%%%%%%%%%%%
\begin{abstract}
Optical microrobots actuated by optical tweezers (OT) are important for cell manipulation and microscale assembly, but their autonomous operation depends on accurate 3D perception. Developing such perception systems is challenging because large-scale, high-quality microscopy datasets are scarce, owing to complex fabrication processes and labor-intensive annotation. Although generative AI offers a promising route for data augmentation, existing generative adversarial network (GAN)-based methods struggle to reproduce key optical characteristics, particularly depth-dependent diffraction and defocus effects.
To address this limitation, we propose Du-FreqNet, a dual-control, frequency-aware diffusion model for physically consistent microscopy image synthesis.
The framework features two independent ControlNet branches to encode microrobot 3D point clouds and depth-specific mesh layers, respectively. Crucially, we introduce an adaptive frequency-domain loss that dynamically reweights high- and low-frequency components based on the distance to the focal plane.
By leveraging differentiable FFT-based supervision, Du-FreqNet captures physically meaningful frequency distributions often missed by pixel-space methods. 
Trained on a limited dataset (e.g., 80 images per pose), our model achieves controllable, depth-dependent image synthesis, improving SSIM by 20.7\% over baselines. 
Extensive experiments demonstrate that Du-FreqNet generalizes effectively to unseen poses and significantly enhances downstream tasks, including 3D pose and depth estimation, thereby facilitating robust closed-loop control in microrobotic systems.
\end{abstract}

\acrodef{AI}[AI]{Artificial Intelligence}
\acrodef{GAN}[GAN]{Generative Adversarial Network}
\acrodef{LDM}[LDM]{Latent Diffusion Model}
\acrodef{MSE}[MSE]{Mean Square Error}
\acrodef{OT}[OT]{Optical Tweezer}
\acrodef{OTF}[OTF]{Optical Transfer Function}
\acrodef{PSF}[PSF]{Point Spread Function}

%%%%%%%%%%%%%%%%%%%%%%%%%%%%%%%%%%%%%%%%%%%%%%%%%%%%%%%%%%%%%%%%%%%%%%%%%%%%%%%%
\section{Introduction}
Optical microrobots actuated by optical tweezers (OTs) have emerged as a promising platform for biomedical manipulation \cite{zhang2020distributed,ali2025optical,wang2025optical}. Through the momentum transfer of light, these systems enable precise and non-invasive handling of microscopic objects, ranging from single cells to synthetic micro-objects~\cite{bustamante2021optical}. However, achieving autonomous operation of optical microrobots for biomedical applications requires robust visual perception for closed-loop control. In particular, effective control depends on real-time estimation of the microrobot’s six-degree-of-freedom (6-DoF) state, namely its three-dimensional (3D) position \((x,y,z)\) and orientation \((\mathrm{roll}, \mathrm{pitch}, \mathrm{yaw})\), from two-dimensional (2D) microscopy images~\cite{wei2025physics}. Although deep learning has substantially advanced perception in macroscopic robotics, progress at the microscale remains limited by the scarcity of sufficiently large and well-annotated microscopy datasets~\cite{wei2025dataset,qiegen2025diffusion,kebaili2023deep,goceri2023medical,groger2025review}.

%Learning-based six-degree-of-freedom (6-DoF) perception for optical microrobots is limited by the scarcity of large, diverse and consistently annotated microscopy datasets \cite{zhang2020data}. 

This challenge is especially pronounced in optical microrobotics, where dataset acquisition is constrained by both experimental burden and the difficulty of obtaining reliable labels \cite{zhang2020data}. Fabricating diverse microrobot geometries is laborious, while establishing ground-truth depth labels, particularly along the optical axis, typically requires specialised hardware such as piezo-stages or confocal reference systems \cite{zhang2022fabrication}. In addition, optical microscopy image formation is governed by diffraction, depth-dependent defocus and spatially varying illumination~\cite{li2022incorporating,ning2023deep,guo2025deep}. These effects create a highly non-linear mapping from the microrobot’s 3D state to its 2D image appearance, limiting the ability of simple augmentation methods to reproduce the visual cues required for robust perception~\cite{kebaili2023deep,goceri2023medical,groger2025review,liu2025state,li2022incorporating}. Together, these constraints motivate the development of controllable and physically informed microscopy image synthesis for optical microrobot perception~\cite{liu2025state,li2022incorporating,qiao2024zero,zhang2025pixel}.

%Recent benchmarks show that pose and depth estimation under optical microscopy is feasible, but that performance depends strongly on data scale and annotation quality~\cite{wei2025dataset}. Related work on detection, tracking and orientation estimation further indicates that reliable perception depends on subtle appearance variations induced by depth and rotation~\cite{choudhary2025three}. 

To address data scarcity, synthetic data generation has emerged as an attractive alternative to large-scale experimental collection. Physics-based simulators can provide geometrically accurate labels, but they often fail to reproduce the noise characteristics, optical aberrations and texture statistics of real microscopy images, resulting in a substantial sim-to-real domain gap \cite{11127419}. Generative adversarial networks (GANs) have been introduced to reduce this gap by translating synthetic sketches into more realistic microscopy images~\cite{zhang2022micro,tan2025physics}. However, although GAN-based methods can improve visual realism, they remain prone to training instability and mode collapse. More fundamentally, because they operate predominantly in the spatial domain and do not explicitly model physics-aware image formation, they often fail to reproduce the diffraction rings and point spread function (PSF) variations that distinguish in-focus from out-of-focus objects in microscopy~\cite{khayatkhoei2022spatial,saad2024survey}. Synthetic images that do not preserve these optical cues are therefore of limited value for downstream tasks, such as pose and depth estimation of microrobots.

Denoising diffusion probabilistic models (DDPMs) and their latent variants have recently surpassed GANs in image quality, training stability and mode coverage~\cite{dhariwal2021diffusion,qiegen2025diffusion,liu2025state}, making them a promising alternative for synthetic data generation in microrobotic microscopy. Applying standard diffusion models to this setting, however, reveals an important limitation. Although Parseval’s theorem implies a formal equivalence between minimising spatial mean-squared error and minimising spectral error~\cite{jiang2021focal}, standard diffusion objectives remain insufficiently sensitive in practice to high-frequency components. This limitation arises from both the spectral bias of neural networks~\cite{rahaman2019spectral,qiao2024zero,zhang2025pixel} and the unequal distribution of energy across frequency bands~\cite{jiang2021focal}. In microscopy, this weakness is especially consequential because axial depth is encoded not only geometrically, but also through depth-dependent changes in image spectral content. Prior diffusion studies in microscopy have focused mainly on super-resolution, structural dataset generation, or image restoration with physical priors~\cite{saguy2025microtubule,eschweiler2024denoising,li2024microscopy,li2022incorporating,guo2025deep,liu2025conditional}. However, these approaches generally do not support explicit conditioning on both the microrobot’s 3D configuration and its focal-plane offset, nor do they directly enforce the depth-dependent frequency characteristics associated with diffraction and defocus.

As optical microrobots move along the axial direction, their microscopy appearance changes systematically with focal state. These changes are not fully captured by geometric representations alone, because defocus also alters the distribution of spectral energy in a depth-dependent manner. Motivated by this observation, we introduce \textbf{Du-FreqNet (Dual-Control Frequency-Aware Network)}, a physics-informed diffusion framework for depth-dependent microscopy of optical microrobots. 
The primary contributions of this paper are summarised as follows:
\begin{enumerate}
    \item \textbf{Dual-Control Architecture}: We introduce a novel diffusion architecture featuring two independent ControlNet branches. The first encodes the microrobot’s 3D point cloud to ensure volumetric structural integrity, while the second encodes depth-stratified mesh projections to explicitly guide the rendering of depth-dependent optical effects.
    \item \textbf{Adaptive Frequency-Domain Loss}: We propose a physics-informed loss function that operates in the Fourier domain. By dynamically reweighting high- and low-frequency components based on the object's distance from the focal plane, this loss forces the model to learn the correct diffraction patterns and defocus blur associated with specific depths.
    \item \textbf{Sim-to-Real Verification}: We demonstrate that training pose and depth estimation networks on Du-FreqNet-generated data significantly outperforms baselines trained on other synthetic data. Our method improves SSIM by 20.7\% and generalises effectively to unseen poses, facilitating robust closed-loop control in microrobotic systems.
\end{enumerate}

% \section{Related Work}
% xxx

\section{Related Work}

\subsection{Data scarcity and synthetic data needs in optical-tweezer microrobotics}
Optical microrobots manipulated under optical tweezers require reliable 6-DoF state estimation from 2D microscopy feedback, yet learning-based perception is fundamentally constrained by the lack of large, diverse, and consistently annotated datasets.
Recent efforts have begun to standardize benchmarking and release task-focused datasets for pose/depth perception under optical microscopy, which highlights both feasibility and the strong data dependence of modern models \cite{wei2025dataset}.
Complementary perception pipelines for optical microrobots (e.g., detection, tracking, and 3-axis orientation estimation) further demonstrate that performance hinges on capturing subtle appearance cues induced by depth and rotation \cite{choudhary2025three}.
In practice, curating such data is expensive due to microrobot fabrication variability, microscope configuration changes, and labor-intensive annotation, while the wider optical-tweezers community has also emphasized the need for more systematic, reusable data practices \cite{halma2025fair}.
These constraints motivate synthetic data augmentation; however, for OT microscopy, naive augmentations often fail because defocus blur, diffraction rings, and PSF variations change nonlinearly with depth and optical settings.
Therefore, controllable and physically consistent microscopy image synthesis is critical for scaling perception models and improving sim-to-real transfer in microrobotic closed-loop control.

\subsection{Physics-consistent synthesis and diffusion models for optical microscopy}
Traditional physics-based image formation models and simulators provide clean geometric labels, but matching real microscope statistics is difficult because realistic PSF variation, diffraction artifacts, and noise processes are hard to reproduce at scale.
In the last three years, diffusion models have emerged as a strong alternative for microscopy image generation and augmentation, offering improved stability and diversity compared with GAN-based pipelines.
For example, diffusion has been used to generate realistic super-resolution microscopy imagery \cite{saguy2025microtubule} and to synthesize fully annotated microscopy datasets by conditioning the generative process on coarse structural sketches or masks \cite{eschweiler2024denoising}.
Beyond pure generation, diffusion has also been adapted to microscopy restoration with explicit physical priors: physics-informed diffusion incorporates light-propagation or PSF-based constraints to reduce artifacts and hallucinations when training data are limited \cite{li2024microscopy}.
Related conditional diffusion frameworks have been explored to recover optically-sectioned content from wide-field inputs, implicitly leveraging depth-dependent degradation \cite{liu2025conditional}.
Despite these advances, most existing microscopy diffusion methods target fluorescence restoration/translation or generic structural synthesis, and they rarely provide (i) explicit depth-controllable generation tailored to OT microrobot appearances or (ii) supervision that directly enforces depth-coupled spectral/PSF characteristics critical for diffraction- and defocus-dominated optical microscopy.
This gap motivates diffusion-based synthesis that is both controllable in depth/pose and physically grounded in the frequency-domain signatures of optical image formation.

\section{Method}
\subsection{Preliminaries and Problem Formulation}

\begin{figure}[!t]
\centering
\captionsetup{font=footnotesize,labelsep=period}
\includegraphics[width=1\columnwidth]{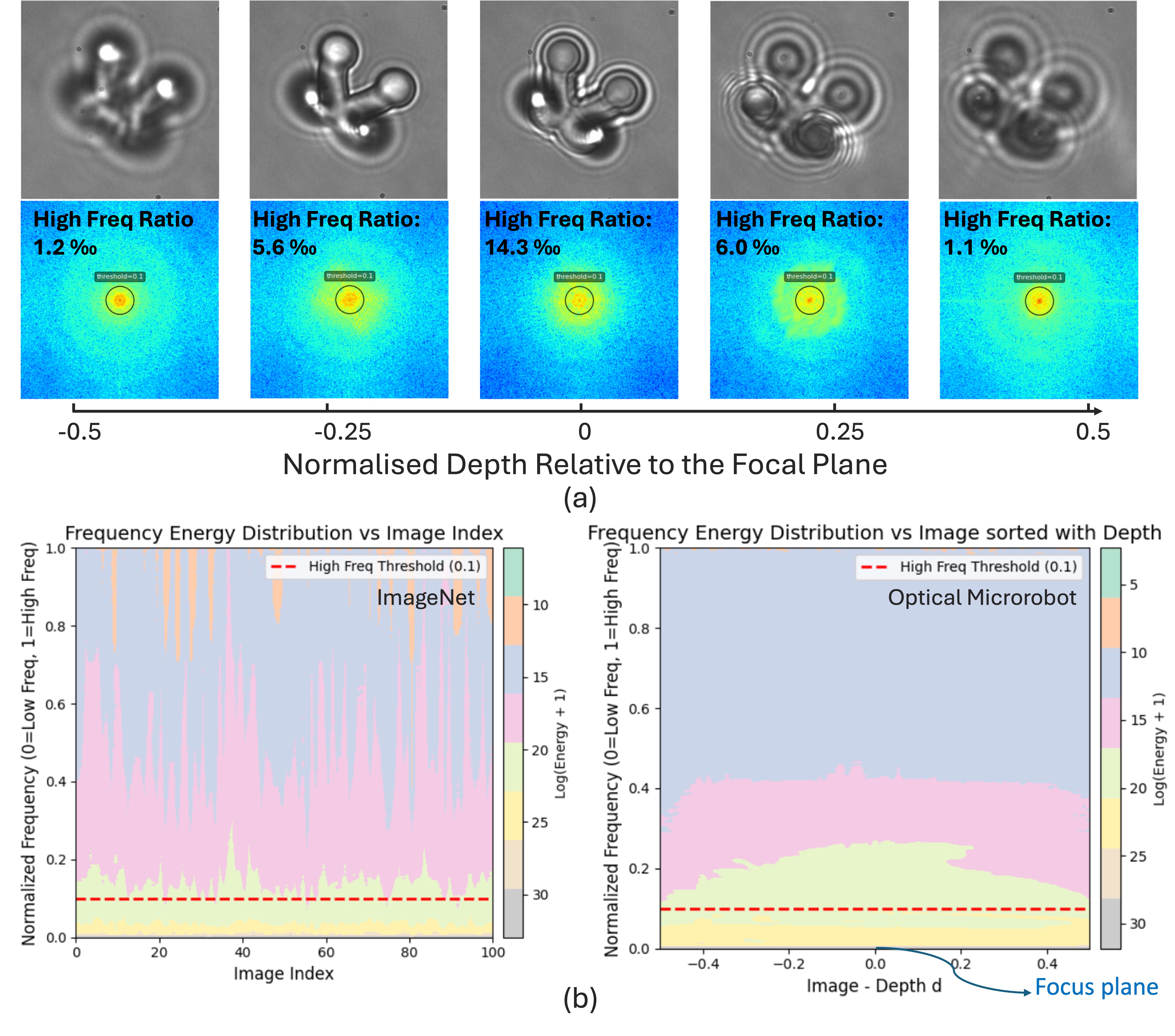}
\caption {%Motivation Analysis. (a) Depth-Frequency Correlation: The high-frequency ratio serves as a proxy for focus. 
Depth–frequency relationship in optical microrobot microscopy. (a) The high-frequency energy ratio as a function of normalised depth, demonstrating its role as a proxy for focal state.
It peaks at the focal plane ($d=0$) and decays by an order of magnitude ($\sim10\times$) in defocused states, confirming that spectral energy is tightly coupled with axial depth.
(b) Distribution Comparison: Unlike the stochastic spectral distribution of natural images (ImageNet, left), our microrobot dataset (right) exhibits a predictable relationship between high-frequency energy and depth relative to the focal plane.
}
\label{fig:motivation}
\vspace{0cm}
\end{figure}

To determine whether axial depth leaves a measurable spectral signature in optical microrobot microscopy, we analysed the frequency content of images acquired at different distances from the focal plane. As shown in Fig.~\ref{fig:motivation}, microrobot images exhibit a clear depth-dependent redistribution of spectral energy. In contrast to natural images, such as those in ImageNet, which show broad and comparatively unstructured frequency distributions (Fig.~\ref{fig:motivation}b, left), our microscopy data display a systematic coupling between focal state and high-frequency content (Fig.~\ref{fig:motivation}b, right).

This trend is quantified in Fig.~\ref{fig:motivation}a using the high-frequency energy ratio as a function of normalised depth. The ratio peaks at the focal plane and decreases by approximately an order of magnitude as the microrobot moves out of focus. This behaviour is consistent with defocus suppressing high spatial frequencies while in-focus images retain sharper diffraction-related structure. Axial depth is therefore encoded not only through projected geometry, but also through a characteristic depth-dependent spectral profile.

These findings suggest that high-fidelity image synthesis requires more than geometric conditioning alone. A successful model must also preserve the spectral cues associated with focus and defocus, motivating the use of physics-informed, frequency-aware supervision in Du-FreqNet.

We aim to synthesize high-fidelity microscopy images $x \in \mathbb{R}^{H \times W \times 3}$ that are physically consistent with a microrobot's 3D geometry and its optical depth relative to the focal plane. 
This is formulated as a conditional generation problem $p(x | \mathcal{P}, \mathcal{M}, d, \tau)$, where $\mathcal{P}$ represents the volumetric point cloud of the microrobot, $\mathcal{M}$ represents the depth-specific mesh cross-section, $d \in [-0.5, 0.5]$ denotes the normalized distance to the focal plane, and $\tau$ is the text prompt.
We leverage \acp{LDM}~\cite{rombach2022high} as our backbone. 
\acp{LDM} operate in a compressed latent space $z = \mathcal{E}(x)$ rather than pixel space to reduce computational complexity. The diffusion process progressively destroys the structure of the latent code $z_0$ by adding Gaussian noise over $T$ timesteps, producing $z_t$. 
The learning objective is to train a denoising autoencoder $\epsilon_\theta$ to predict the added noise $\epsilon$ given the current noisy state $z_t$, the timestep $t$, and the set of conditions $C = \{\mathcal{P}, \mathcal{M}, d, \tau\}$:
$$\mathcal{L}_{diff} = \mathbb{E}_{z_0, t, C, \epsilon \sim \mathcal{N}(0,1)} \left[ \| \epsilon - \epsilon_\theta(z_t, t, C) \|_2^2 \right].$$

\subsection{Overview of Du-FreqNet Architecture}
\begin{figure*}[!th]
\captionsetup{font=footnotesize,labelsep=period}
\centering \includegraphics[width=0.95\linewidth]{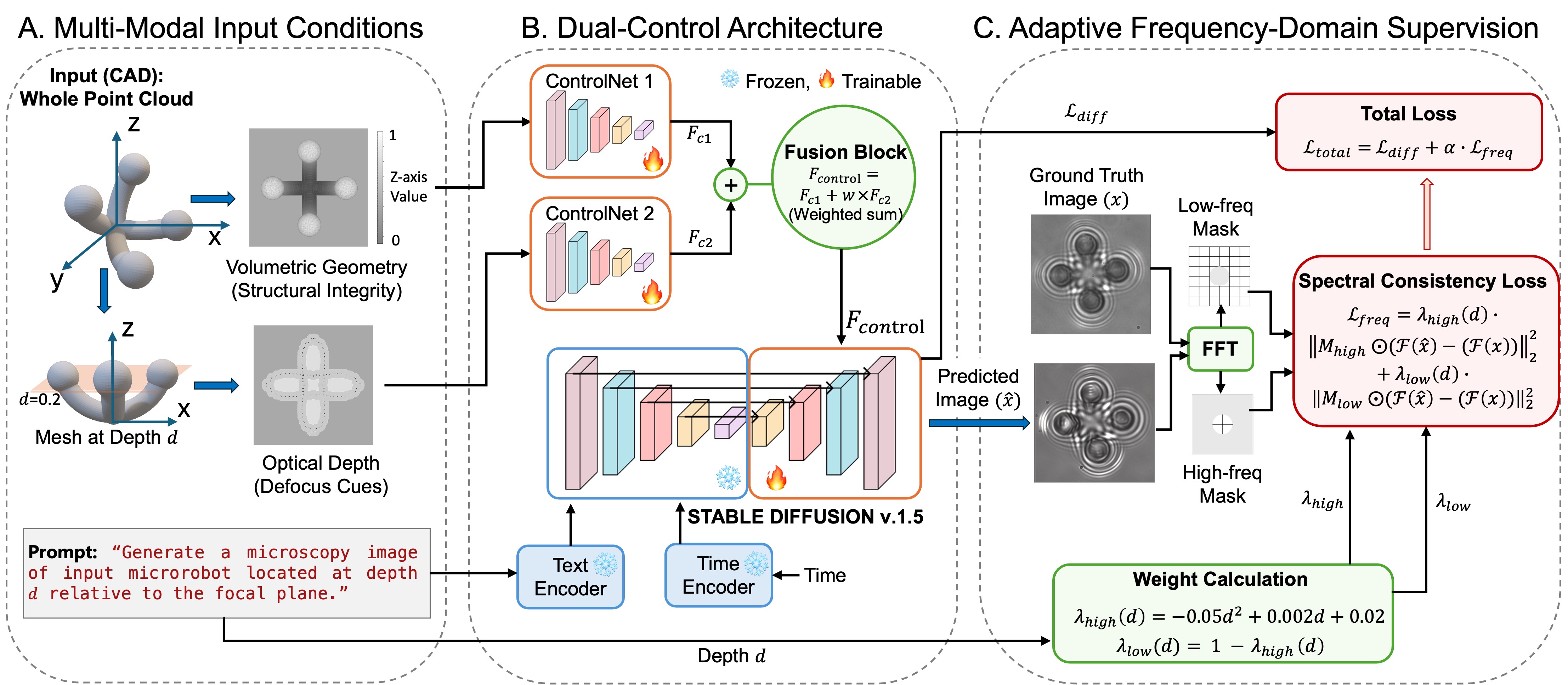}
\caption {Overview of the proposed Du-FreqNet framework. 
(A) Multi-Modal Input Conditions: The model utilises two distinct geometric priors: volumetric point clouds to ensure structural integrity and depth-specific mesh slices to encode optical defocus cues. 
(B) Dual-Control Architecture: These conditions are processed by parallel ControlNet branches and fused via a weighted sum before injection into a Stable Diffusion backbone. 
(C) Adaptive Frequency-Domain Supervision: To enforce physical consistency, an adaptive spectral loss dynamically reweights high- and low-frequency components (via FFT) based on the microrobot's depth $d$, explicitly modelling the depth-dependent decay of diffraction patterns.
}
\label{fig:framework}
\end{figure*}

To address the limitations of standard diffusion models in handling optical physics, we propose Du-FreqNet (Dual-Control Frequency-Aware Network). As illustrated in Fig.~\ref{fig:framework}, the architecture extends the standard Stable Diffusion (SD) framework with two distinct mechanisms: 
\begin{enumerate}
    \item Dual-Control Input: We employ two parallel ControlNet branches to decouple the volumetric geometry of the robot from the optical characteristics of the imaging plane. 
    The outputs of these branches are fused via weighted element-wise addition before being injected into the unlocked SD U-Net backbone.
    \item Frequency-Aware Supervision: Unlike standard ControlNets that rely solely on spatial reconstruction losses, Du-FreqNet introduces a Spectral Consistency Loss ($\mathcal{L}_{freq}$). 
    This module dynamically reweights the supervision of high- and low-frequency components based on the depth $d$ extracted from the text prompt, ensuring the generated images respect the physics of diffraction and defocus.
\end{enumerate}

\subsection{Dual-Control Mechanism}
Microrobot microscopy images are complex because the appearance depends on both the 3D shape of the object and the 2D slice being imaged. 
A single control condition is insufficient to capture this duality. 
We therefore design two specialised input streams:
\begin{enumerate}
    \item Volumetric Geometry Branch (Control Source 1): This branch ensures the structural integrity of the generated robot. 
    The input is a projection of the whole point cloud mesh structure ($\mathcal{P}$), where the intensity encodes the global $z$-coordinates of the robot. 
    This condition provides a "Volumetric Prior" preventing the model from distorting the robot's morphology regardless of the focal settings.
    \item Optical Depth Branch (Control Source 2): This branch explicitly models the intersection of the focal plane with the robot. 
    The input is a rendered map of points at a certain depth ($\mathcal{M}$), representing the specific cross-section of the robot currently positioned at the focal plane. 
    % As shown in the "Target Image Alignment", this signal correlates strongly with the sharpest regions in the ground truth image (high Laplacian of Gaussian response). 
    By isolating this slice, the network learns where to render sharp textures versus diffraction rings.
    These two control signals are processed by independent ControlNet encoders to extract features. 
    To allow for flexible modulation between geometric structural guidance and optical depth cues, we aggregate these features via weighted sum:
    $$F_{control} = F_{c1}(z_t, \mathcal{P}) + w F_{c2}(z_t, \mathcal{M}).$$ 
    This merged feature $F_{control}$ is then added to the skip connections of the SD U-Net. In our experiments, we set $w$= 0.3.
\end{enumerate}

\subsection{Adaptive Frequency-Domain Supervision}
To theoretically justify our spectral objective, we first consider the limitations of standard pixel-space reconstruction from a signal processing perspective. Standard diffusion models typically minimise the \ac{MSE}, $\mathcal{L}_{pix} = \| x - \hat{x} \|_2^2$. 
According to Parseval’s Theorem~\cite{jiang2021focal}, the energy of the error in the spatial domain is preserved in the frequency domain:
$$\| x - \hat{x} \|_2^2 = \frac{1}{HW} \| \mathcal{F}(x) - \mathcal{F}(\hat{x}) \|_2^2.$$
This unitary invariance implies that standard \ac{MSE} assigns uniform importance to all frequency components~\cite{jiang2021focal}. 
However, in diffraction-limited optical microscopy, the semantic information regarding "focus" is disproportionately concentrated in high-frequency bands, which constitute a minute fraction of the total spectral energy. 
Consequently, a standard \ac{MSE} loss is dominated by low-frequency structural errors and often fails to capture the subtle high-frequency diffraction patterns that define the focal state.

From an optical physics standpoint, this depth-dependency is governed by the system's \ac{OTF}, which acts as a depth-varying low-pass filter, where the contrast transfer at high spatial frequencies decays significantly as the defocus distance $|d|$ increases~\cite{prakash2025resolution}. 
While the exact analytical decay follows a complex Bessel function distribution (often approximated by a $\text{sinc}^2$ function), our empirical analysis of the microrobot dataset (Fig.~\ref{fig:motivation}) indicates that within the operational workspace, this spectral decay can be robustly approximated by a quadratic function.

Therefore, we introduce an Adaptive Frequency Loss ($\mathcal{L}_{freq}$) that breaks the uniform weighting of Parseval's theorem. 
By dynamically reweighting spectral components based on the physics of the \ac{OTF}, we force the network to explicitly minimise the error in high-frequency bands only when the physical depth suggests they should exist. 
After analysing the optical microrobot data, we define the depth-dependent weighting coefficients, high-frequency weight $\lambda_{high}(d)$ and low-frequency weight $\lambda_{low}(d)$ as follows:
% % A core insight of this work is that the "realism" of microscopy data is encoded in the frequency domain. As established in our motivation, the ratio of high-frequency energy decays predictably as the object moves away from the focal plane. Standard pixel-wise losses (MSE) struggle to capture these spectral nuances.We introduce an Adaptive Frequency Loss that operates in the Fourier domain. The depth $d$ is parsed directly from the text prompt (e.g., "Depth to the focal plane: 0.25").
% \begin{enumerate}
%     \item Dynamic Frequency Weighting: We empirically model the relationship between depth and high-frequency importance using a quadratic decay function derived from dataset statistics. We define the high-frequency weight $\lambda_{high}(d)$ and low-frequency weight $\lambda_{low}(d)$ as:
$$\lambda_{high}(d) = -0.05 d^2 + 0.002 d + 0.02,$$
$$\lambda_{low}(d) = 1 - \lambda_{high}(d).$$
This formulation assigns maximum weight to high frequencies when the robot is at the focal plane ($d \approx 0$) and significantly reduces it as $|d| \to 0.5$ (defocused).

To apply these weights, we generate soft masks for the 2D Fast Fourier Transform (FFT) spectrum. Let $\mathcal{F}(\cdot)$ denote the 2D FFT operation shifted to center low frequencies. We define high-frequency ($M_{high}$) and low-frequency ($M_{low}$) masks based on the normalized radial distance $r$ from the DC component, using a sigmoid transition at a threshold $\tau_{freq}=0.1$:
$$M_{high}(r) = \sigma(s \cdot (r - \tau_{freq})),$$
$$\quad M_{low}(r) = 1 - M_{high}(r),$$
where $s$ controls the steepness of the transition.
The final frequency loss measures the weighted spectral distance between the generated image $\hat{x}$ and the ground truth $x$:
% \begin{equation}
% \begin{aligned}
% \mathcal{L}_{freq} = &\lambda_{high}(d) \cdot \| M_{high} \odot (\mathcal{F}(\hat{x}) - \mathcal{F}(x)) \|_2^2 +\\ &\lambda_{low}(d) \cdot \| M_{low} \odot (\mathcal{F}(\hat{x}) - \mathcal{F}(x)) \|_2^2, \nonumber
% \end{aligned}
% \end{equation}
\begin{equation}
\begin{aligned}
\mathcal{L}_{freq} = \textbf{1}_{[t < T]} \cdot \Big( 
    &\lambda_{high}(d) \cdot \| M_{high} \odot (\mathcal{F}(\hat{x}_0) - \mathcal{F}(x_0)) \|_2^2 \\
    + &\lambda_{low}(d) \cdot \| M_{low} \odot (\mathcal{F}(\hat{x}_0) - \mathcal{F}(x_0)) \|_2^2 
\Big),
\end{aligned}
\end{equation}
where $\odot$ denotes the Hadamard product and $\textbf{1}_{[t < T]}$ is an indicator function that activates the loss only when the diffusion timestep $t$ is below a threshold $T$ (empirically set to 500). This selective application is motivated by the observation that predictions of $\hat{x}_0$ at early timesteps (large $t$) are predominantly noisy and unreliable for meaningful frequency analysis.
This loss forces the model to prioritize sharp edges and textures only when the physics dictates they should exist (i.e., near the focal plane) while focusing on the structural shape (low frequency) in defocused regions.

\subsection{Total Training Objective}
The full training objective combines the latent diffusion objective with the spectral supervision. 
Since the frequency loss is computed in the pixel space (requiring decoding of latents), we apply it with a regularization weight $\alpha$:
$$\mathcal{L}_{total} = \mathcal{L}_{diff} + \alpha \cdot \mathcal{L}_{freq}.$$
In our experiments, we set $\alpha=0.001$. 
This composite objective ensures that Du-FreqNet generates images that are both semantically correct (controlled by $\mathcal{L}_{diff}$ and the dual prompts) and physically consistent with optical laws (controlled by $\mathcal{L}_{freq}$).

\section{Experiments and Results}
\subsection{Dataset}
The experimental dataset was constructed using a holographic Optical Tweezers (OT) manipulation system (Elliot Scientific, UK) integrated with a high-precision nanopositioner (Mad City Labs Inc., USA). 
The target microrobots were fabricated using Two-Photon Polymerization (TPP) on an IP-L photoresist via a Nanoscribe 3D printer (Nanoscribe GmbH, Germany). 
Visual feedback was captured using a CCD-coupled optical microscope, recording raw frames at a resolution of $678 \times 488$ pixels.

To establish high-precision Ground Truth labels for 3D perception, the microrobot substrates were manipulated using piezoelectric stages.
The depth, which is the trajectories of the robot along the optical axis were generated by moving the piezoelectric substrate, providing continuous depth labels.
Pose refers to the robot's out-of-plane rotations, which were achieved via a piezoelectric drive operating in discrete modes. 
We define the pose state based on Pitch ($P$, rotation around the $x$-axis) and Roll ($R$, rotation around the $y$-axis).
The dataset spans a discrete pose space with an angular resolution of $10^{\circ}$, resulting in a total of 35 distinct pose classes. Further details regarding the data collection protocol are available in~\cite{wei2025dataset}.

Prior to training, raw microscopy images underwent a standardised preprocessing sequence to mitigate sensor noise and illumination variations. 
Images were first converted to grayscale and denoised using a pixel-wise adaptive low-pass Wiener filter. 
To isolate the microrobot, we applied automatic threshold binarization followed by Canny edge detection. 
Based on the computed centroid of the microrobot, a consistent Region of Interest (ROI) was cropped and normalised for all subsequent analyses.

\subsection{Experimental Setup}
We utilised the data of 100 images per pose class for model development. 
To evaluate both reconstruction quality and generalisation capability, the dataset was partitioned into two distinct subsets:
1) Seen configurations: 
For 31 of the 35 collected pose classes, we employed a stratified random split of 80\% for training, 10\% for validation, and 10\% for testing. 
This subset evaluates the model's ability to synthesise depth-dependent variations for known geometries.
2) Unseen configurations: We explicitly excluded 4 pose classes from the training set to evaluate zero-shot generalisation. 
% These unseen poses were categorised into two groups to test structural robustness:
% symmetric poses: $P20^{\circ}\_R20^{\circ}$ and $P50^{\circ}\_R50^{\circ}$; 
% asymmetric poses: $P10^{\circ}\_R40^{\circ}$ and $P30^{\circ}\_R70^{\circ}$.

\textbf{Du-FreqNet Training:} 
All experiments were implemented using PyTorch 1.10 within a Python 3.8 environment. 
All experiments were conducted on a single NVIDIA A100 GPU (80 GB) running CUDA 12.0. 
Following the standard ControlNet paradigm~\cite{zhang2023adding}, we initialised the backbone using pre-trained weights from Stable Diffusion v1.5. 
Input images were resized to a resolution of $512 \times 512$ pixels. 
The network was optimised using the AdamW optimiser with a fixed learning rate of $1 \times 10^{-5}$ and a batch size of 32. 
For inference and training sampling, we employed the DDIM sampler with a linear noise schedule. 
To ensure convergence and prevent overfitting, training was governed by an early stopping mechanism with a patience of 20 epochs, capped at a maximum of 300 epochs.

\begin{table*}[!t]
\centering
\captionsetup{font=footnotesize,labelsep=period}
\caption{Quantitative Comparison of Image Synthesis Quality. Best results are bolded. Our Du-FreqNet achieves superior performance across structural (SSIM, PSNR) and perceptual (FID, LPIPS) metrics compared to baselines.}
\renewcommand{\arraystretch}{1.2}
\begin{tabular}{c:ccccc}
\hline\hline
Model/Value   & SSIM ($\uparrow$)   & PSNR ($\uparrow$)   & MSE (e-02) ($\downarrow$) &  LPIPS $\downarrow$ & FID $\downarrow$
\\ \hline
Physics Rendering  & 0.632 & 15.07  &  3.330 & 0.504 & 229.342 \\ \hline
Pix2Pix GAN &  0.613  &  16.65 &  2.210  & 0.334  & 150.558 \\ \hline
Physics Rendering + GAN  & 0.641  & 16.88  & 2.090 & 0.333 & 135.678 \\ \hline
ControlNet  & 0.685 & 19.32&1.308 & 0.222& 54.060 \\ \hline
\textbf{Du-FreqNet}  & \textbf{0.827}  & \textbf{25.12}  & \textbf{0.548}  & \textbf{0.085} & \textbf{17.444}\\ 
\hline\hline
\end{tabular}
\label{table:comparison}
\end{table*}

\begin{figure*}[!th]
\captionsetup{font=footnotesize,labelsep=period}
\centering \includegraphics[width=0.85\linewidth]{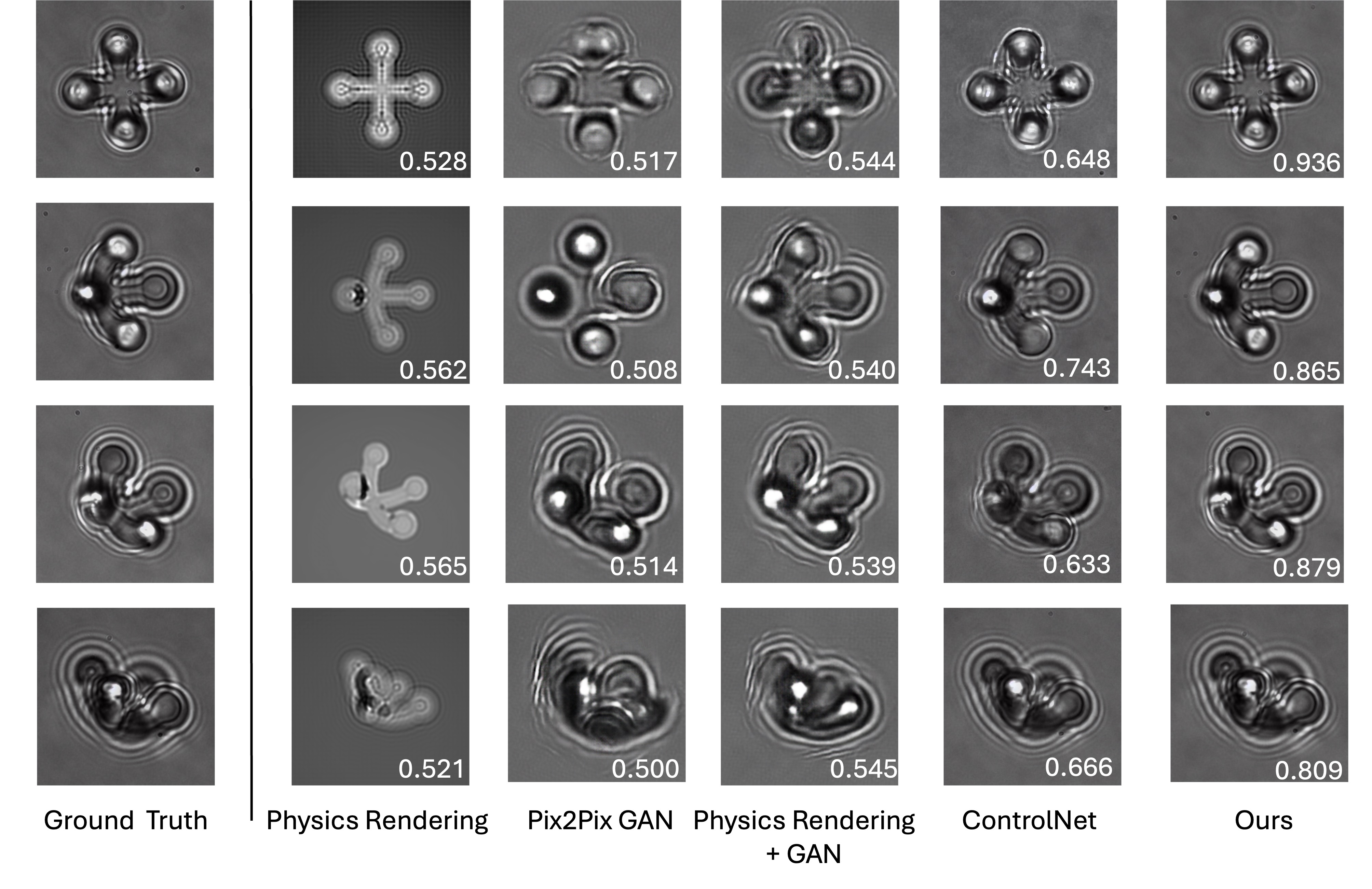}
\vspace{-0.3cm}
\caption {Visualisation of image synthesis results. White numbers indicate the SSIM value relative to the Ground Truth. Our method preserves high-frequency diffraction patterns and object boundaries better than baselines, achieving superior structural fidelity.
}
\label{fig:result1}
\end{figure*}

\textbf{Baselines Model Training:}
To evaluate the efficacy of our method, we benchmarked against the physics rendering~\cite{tan2025physics} and several GAN-based methods~\cite{isola2017image}. 
For a fair comparison, all baselines were trained on identical training/validation splits and utilised the same depth-related conditioning as our method. 
For the Pix2Pix GAN, we use Control Source 2 as input and for the  Physics Rendering + GAN model, we use the depth-related physics rendering image as input.
Baseline GANs were trained at a standard resolution of $256 \times 256$ pixels with a batch size of 32. 
The training objective combined a vanilla adversarial loss with an $L_1$ reconstruction term weighted by $\lambda=100$ to ensure structural fidelity. We adopted a standard learning rate schedule: a constant learning rate of $2 \times 10^{-4}$ for the initial 200 epochs, followed by a linear decay to zero over the final 100 epochs, totaling 300 epochs.

\subsection{Evaluation Metric}
We evaluate the fidelity of generated images using five complementary metrics:
we employ the SSIM to assess the preservation of high-frequency structural details. 
We quantify absolute reconstruction accuracy using \ac{MSE} and PSNR, measuring the pixel-wise deviation.
% between the generated microrobot images ($\hat{x}$) and their ground truth counterparts ($x$).
% SSIM evaluates luminance, contrast, and structure, which are critical for resolving the microrobot's contact geometry and diffraction patterns.
To capture perceptual similarity that aligns with human vision, we compute the LPIPS distance. 
Additionally, we use the FID to measure the distributional distance between the feature representations of the real and synthetic datasets, evaluating the overall realism and diversity of the generated samples.

Beyond visual quality, we assess the practical utility of our synthetic data for downstream robotic perception tasks. 
We train a ResNet50-based regressor and classifier on the generated data and evaluate their performance on real held-out test sets.
Performance is measured across the microrobot's Pitch, Roll, and Depth.
For discrete pose classification, we report Accuracy (Acc). 
For continuous depth regression, we utilise \ac{MSE}, RMSE and MAE. 
These metrics quantify the extent to which the synthetic images preserve the precise geometric information required for robust closed-loop robotic control.

\begin{table*}[!t]
\centering
\captionsetup{font=footnotesize,labelsep=period}
\caption{Performance comparison of different models on unseen poses. The test set is categorized into sym (Unseen Symmetric poses) and asym (Unseen Asymmetric poses). Du-FreqNet consistently demonstrates superior zero-shot generalization capabilities. Best results in each column are depicted in boldface.}
\renewcommand{\arraystretch}{1.2}
\begin{tabular}{c:c:ccccc}
\hline\hline
Model/Value  & Poses & SSIM ($\uparrow$)   & PSNR ($\uparrow$)   & MSE (e-02) ($\downarrow$) &  LPIPS $\downarrow$ & FID $\downarrow$ \\ \hline
\multirow{2}{*}{Physics Rendering + GAN} 
& sym & 0.616 & 16.48  & 2.280  & 0.351 &  183.950\\ 
& asym  & 0.628 &  16.99 &  2.021 & 0.332 & 162.651 \\ \hline
\multirow{2}{*}{ControlNet}
& sym  & 0.659  & 18.42  & 1.528  & 0.246 & 85.924 \\ 
& asym&  0.659 &  18.69 & 1.470  & 0.241 & 71.308 \\ \hline
\multirow{2}{*}{\textbf{Du-FreqNet}}
& sym  & \textbf{0.686} & \textbf{18.77}  & \textbf{1.412}  & \textbf{0.193} & \textbf{54.961} \\ 
& asym & \textbf{0.712}  & \textbf{20.17} & \textbf{1.067}  & \textbf{0.176} & \textbf{44.559} \\ 
\hline\hline
\end{tabular}
\label{table:generalise}
\end{table*}

\subsection{Results}
\subsubsection{Quantitative Evaluation of Image Synthesis}
Quantitative comparisons in Table~\ref{table:comparison} demonstrate that Du-FreqNet significantly outperforms all baselines across all metrics. 
Our model achieves the highest SSIM and PSNR scores (20.7\% and 30\% higher than the best baseline, respectively), confirming superior preservation of structural geometry and luminance. 
Furthermore, the substantially lower FID and LPIPS metrics indicate that our generated images align much more closely with real-world distributions and human perceptual standards than GAN-based alternatives.

Qualitative inspection (Fig.~\ref{fig:result1}) corroborates these findings. While GANs exhibit substantial blur and boundary noise, Du-FreqNet produces sharp, physically accurate diffraction patterns. We attribute this to two key factors:
1) Iterative Refinement: Unlike single-step GAN generators that struggle with the semantic gap between depth maps and optical textures, our diffusion-based approach gradually refines high-frequency details, ensuring sharper edges;
2) Background Consistency: By leveraging Stable Diffusion's robust pre-trained priors, Du-FreqNet maintains global spatial coherence, avoiding the severe background deformation often observed in GAN baselines where the generator over-prioritizes the foreground.

\subsubsection{Generalizability to Unseen Configurations}
% 在这个实验中我们选取了两个最好的baseline和我们的Du-FreqNet在\emph{unseen poses}进行图片生成测试。我们选取的\emph{unseen poses}分为两类，一类是Symmetry poses，即robot的pitch 和row值相同 ("P20_R20", "P50_R50"),在Tablue II中用sym表示；一类是Asymmetry poses("P10_R40", "P30_R70"})，即robot的pitch 和row值不同, 在Tablue II中用nsym表示。实验结果证明我们的Du-FreqNet在每种unseen poses上的生成图片质量都高于baseline，且Du-FreqNet在unseen poses上的性能高于ControlNet在seen poses上的结果，证明了我们的模型对unseen poses的泛化性能；另外三个模型在Asymmetry unseen poses上的结果都高于Symmetry unseen poses的，说明在pitch和row的角度不同时，生成图片更xxx。

\begin{figure}[!t]
\captionsetup{font=footnotesize,labelsep=period}
\centering \includegraphics[width=0.95\linewidth]{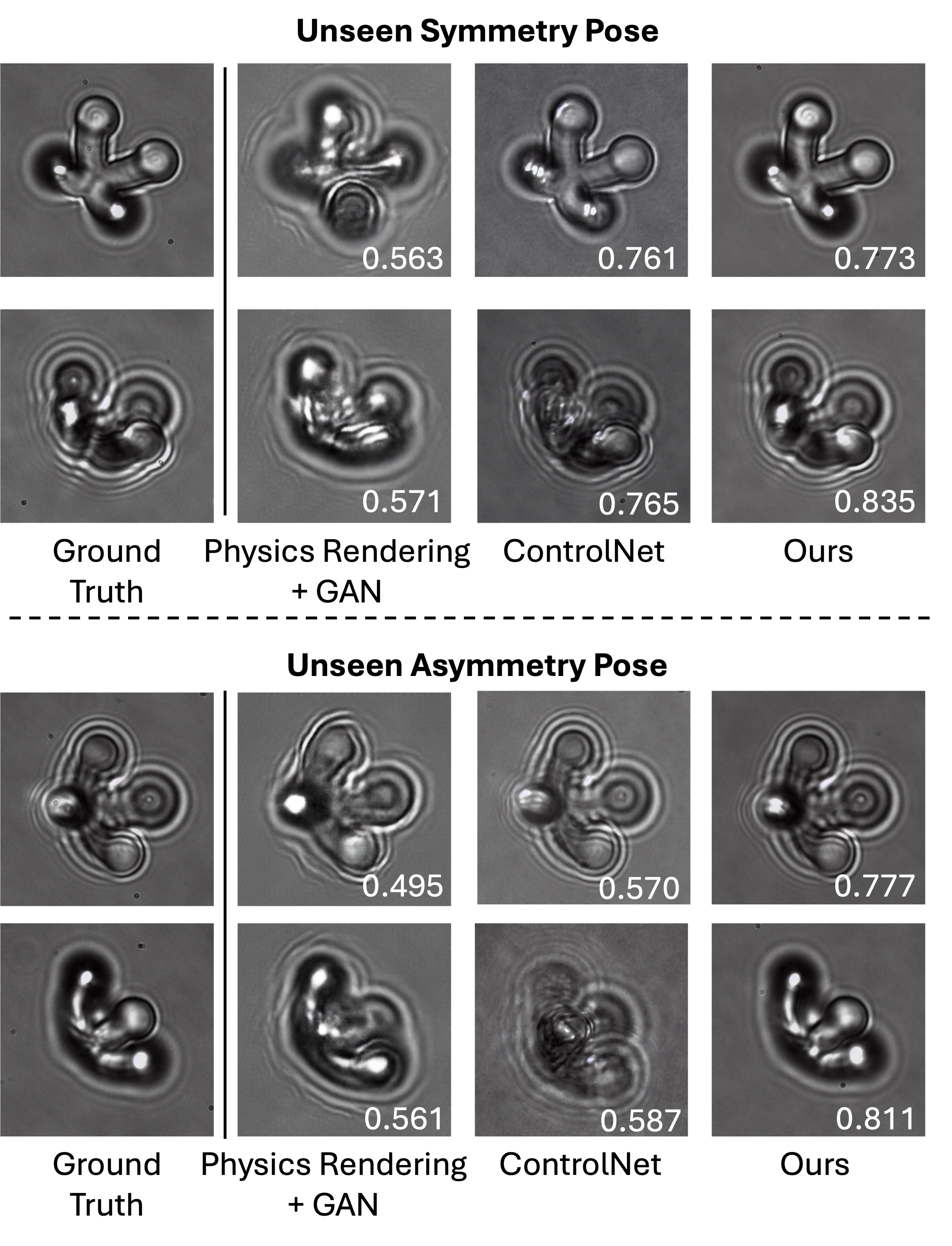}
\caption {Qualitative generalization results on unseen poses.  
The numbers indicate the SSIM value. 
While baselines struggle with structural distortion or blur on novel views, Du-FreqNet successfully generalises, synthesising accurate diffraction patterns and maintaining geometric integrity.
}
\label{fig:sym_asym}
\end{figure}

To evaluate the zero-shot generalization capability of our framework, we conducted experiments on a held-out set of microrobot poses that were explicitly excluded from the training data. 
We benchmark Du-FreqNet against the two best baselines. 
The unseen configurations are categorised into two subsets: 1) Symmetric Poses (Sym): Configurations where the pitch and roll angles are identical (e.g., $P20\_R20$, $P50\_R50$);
2) Asymmetric Poses (Asym): Configurations with distinct pitch and roll angles (e.g., $P10\_R40$, $P30\_R70$).

\textbf{Quantitative Generalization Analysis}:
The quantitative results are summarised in Table~\ref{table:generalise}. 
Du-FreqNet demonstrates superior generalisation across both symmetric and asymmetric categories, consistently outperforming the baselines in all five metrics. In the "Symmetric" category, Du-FreqNet improves SSIM by +4.1\% ($0.686$ vs. $0.659$) and reduces FID by 36.0\% ($54.961$ vs. $85.924$) compared to ControlNet.
Remarkably, our model's performance on unseen asymmetric poses ($0.712$ SSIM, $20.17$ PSNR) surpasses the performance of the standard ControlNet trained on seen poses ($0.685$ SSIM, $19.32$ PSNR, refer to Table~\ref{table:comparison}). 
This suggests that Du-FreqNet learns a robust, physics-compliant underlying representation of the microrobot's optical properties rather than simply memorizing training views.

\textbf{Impact of Pose Symmetry}:
We observe a consistent trend across all models where performance on asymmetric poses is higher than on symmetric poses (e.g., Du-FreqNet SSIM: $0.712$ Asym vs. $0.686$ Sym). 
This can be explained by the fact that asymmetric poses project more distinct geometric features onto the 2D plane, thereby reducing the self-occlusion ambiguity often present in symmetric views. 
This richer spatial information allows the network to render sharper, more discriminative details.

\begin{table*}[h]
\centering
\captionsetup{font=footnotesize,labelsep=period}
\caption{Sim-to-Real verification results on downstream perception tasks. We compare the performance of a ResNet-50 estimator trained on Real, Mixed, and purely Synthetic datasets. The minimal performance degradation (e.g., only 2\% and 0.2\% drop in Pitch and Roll accuracy) confirms the high geometric fidelity and practical utility of our synthetic data.
}
\renewcommand{\arraystretch}{1.2}
% \resizebox{\linewidth}{!}{
\begin{tabular}{c:c|c|c|c|c}
\hline\hline
\multirow{2}{*}{Modality} 
& Pitch & Roll &  \multicolumn{3}{c}{Depth} \\
\cmidrule(lr){2-6}
& Acc. $\uparrow$ & Acc. $\uparrow$ &MSE $\downarrow$ & RMSE $\downarrow$ & MAE $\downarrow$  \\
\hline
Real Dataset &0.995 ± 0.005 &0.988 ± 0.008 &0.054 ± 0.007& 0.233 ± 0.014 &0.184 ± 0.013  \\
\hline
Mixed Dataset & 0.982 ± 0.013& 0.986 ± 0.009&0.057 ± 0.006 &0.238 ± 0.015&0.189 ± 0.015\\
\hline
Synthetic Dataset &0.971 ± 0.015 & 0.986 ± 0.007&0.060 ± 0.007 &0.246 ± 0.015 &0.196 ± 0.015  \\
\hline\hline
\end{tabular}%}
\label{tab:pose_estimation}
\vspace{-0.2cm}
\end{table*}

\textbf{Qualitative Inspection}:
Visual comparisons in Fig.~\ref{fig:sym_asym} corroborate these findings. 
For unseen poses, the baseline (Physics Rendering + GAN) suffers from severe structural collapse (e.g., missing robot arms). While ControlNet captures the general shape, it fails to synthesise the correct diffraction rings associated with the novel depths. In contrast, Du-FreqNet generalises the physical laws of diffraction, producing structurally complete robots with depth-consistent optical interference patterns, even for viewing angles it has never encountered.

\subsection{Downstream Task: Sim-to-Real Verification}

To assess the practical utility of our synthesised data for robotic perception, we fine-tuned a pre-trained ResNet-50 for multi-task prediction. 
The network simultaneously predicts discrete microrobot poses (Pitch and Roll classification via CrossEntropyLoss) and continuous axial position (Depth regression via MSELoss), optimized using a summed multi-task objective. 
We benchmarked three training regimes: Real Dataset (ground truth), Synthetic Dataset (ours), and a Mixed Dataset (50\% Real, 50\% Synthetic), each containing 450 images across 31 poses. Experiments were conducted using 5-fold cross-validation with 30 training epochs per fold.

Table~\ref{tab:pose_estimation} demonstrates that our synthetic data effectively bridges the Sim-to-Real gap. 
Training purely on synthetic data results in a marginal accuracy drop of only 2.4\% for Pitch and 0.2\% for Roll compared to the Real baseline. Notably, the Mixed regime achieves Roll accuracy equivalent to the Synthetic baseline, while Depth regression metrics (MSE, RMSE, MAE) remain comparable across all regimes. 
These results confirm that Du-FreqNet preserves critical high-fidelity geometric information, making it a viable substitute for expensive real-world data collection in training perception systems.

\subsection{Ablation Studies}
\subsubsection{Influence of $w$}
% 为了测试两个control条件的重要性，我们在Dual ControlNet（还没有引入frequency loss）上测试了control2的权重 w \in (0, 1] = 0，0.2，0.3, 0.4， 0.6， 0.8， 1.0。我们发现in table IV，control2的加入可以很大幅度的提升生成图片的质量（SSIM +16%）, 但是一味的增加control2 depth slice的权重并不意味着效果的提升 (w=1.0时的SSIM值和w=0时几乎相同)。当w=0.3时，生成图片的质量最好。这说明在图片生成的过程中，control1（whole point cloud）占重要作用，control2（mesh at certaion depth）占辅助作用。因此，我们选取w=0.3为我们模型中control2条件的权重。
\begin{table}[!t]
\centering
\captionsetup{font=footnotesize,labelsep=period}
\caption{Influence of the fusion weight $w$. We vary the weight of the auxiliary depth-control branch (Control 2) from $0.0$ to $1.0$. The results indicate that while depth cues are essential, they should function as a supplementary guide ($w=0.3$) rather than the dominant signal. Best results are bolded.}
\renewcommand{\arraystretch}{1.2}
\resizebox{\linewidth}{!}{
\begin{tabular}{|c:ccccc|}
\hline
$w$/Value   & SSIM ($\uparrow$)   & PSNR ($\uparrow$)   & MSE (e-02) ($\downarrow$) &  LPIPS $\downarrow$ & FID $\downarrow$
\\ \hline
0.0  & 0.685  & 19.32  &1.308   & 0.222 & 54.060 \\ \hline
0.2  & 0.755  & 21.51  & 0.928  & 0.138 & 24.397 \\ \hline
0.3 &  \textbf{0.795} & \textbf{23.42}  & \textbf{0.709}  & \textbf{0.104} & \textbf{18.573} \\ \hline
0.4  & 0.750  &  21.33 & 0.965  & 0.145 & 25.096 \\ \hline
0.6  & 0.733  & 20.73  & 1.060  & 0.159 & 28.544 \\ \hline
0.8  & 0.742  & 21.14  & 0.978 & 0.150 & 26.914 \\ \hline
1.0  & 0.694  & 19.09  & 1.410  & 0.210 & 31.201  \\ 
\hline
\end{tabular}}
\label{table:w}
\end{table}
To quantify the synergy between the volumetric geometry (Control 1) and depth-specific optical cues (Control 2), we analysed the sensitivity of the model to the fusion weight $w$ assigned to the second control branch. 
We evaluated the Dual-Control architecture (prior to adding frequency loss) by varying $w$ across the range $[0, 1.0]$, where $w=0$ represents a single-branch baseline using only the global point cloud.
The results in Table~\ref{table:w} reveal a distinct performance trajectory. 
Integrating the depth-slice condition significantly enhances generation fidelity, yielding a peak 16.0\% improvement in SSIM ($0.685 \to 0.795$) at $w=0.3$. This confirms that explicit depth information is critical for resolving optical features that the point cloud alone cannot define.
However, we observe that indiscriminately increasing the weight of the depth slice is detrimental; at $w=1.0$, the performance regresses ($0.694$), becoming statistically indistinguishable from the single-branch baseline ($w=0$). 
This implies a hierarchical functional role: the Whole Point Cloud (Control 1) serves as the primary driver for global structural integrity, while the Mesh at Depth (Control 2) serves an auxiliary role, fine-tuning local diffraction details. 
When $w$ is too large, the model overfits to the 2D planar slice and loses the volumetric context required for consistent 3D reconstruction. Consequently, we empirically set $w=0.3$ as the optimal configuration for our framework.

\subsubsection{Influence of $\alpha$}
% follow by上一个section的实验，我们选取最好的Dual ControlNet参数进一步加入frequency loss进行Du-FreqNet的实验，来测试Frequency loss的权重 \alpha 对生成图片的影响。我们测试了 \alpha = 0.0001, 0.001, 0.01, 0.1, 0.5, 1.0。实验结果in Table V，我们发现frequency loss的加入使得生成图片的质量有了提升（SSIM从0.795增加到了0.827）， 但是\alpha量级的变化（从0.0001到0.1）并没有对结果产生大的影响。最终，我们选取\alpha=0.001为Du-FreqNet的模型参数。
\begin{table}[!t]
\centering
\captionsetup{font=footnotesize,labelsep=period}
\caption{Influence of the spectral loss weight $\alpha$ (with fixed $w=0.3$). We evaluate the impact of the frequency loss magnitude. The inclusion of frequency supervision consistently improves image quality compared to the baseline Dual-Control model (Table~\ref{table:w}), with the model showing high robustness to parameter variations. Best results are bolded.}
\renewcommand{\arraystretch}{1.2}
\resizebox{\linewidth}{!}{
\begin{tabular}{|c:ccccc|}
\hline
$\alpha$/Value   & SSIM ($\uparrow$)   & PSNR ($\uparrow$)   & MSE (e-02) ($\downarrow$) &  LPIPS $\downarrow$ & FID $\downarrow$
\\ \hline
0.0001  & 0.825  & 24.90  & 0.577  & 0.089 & 17.259 \\ \hline
0.001   & \textbf{0.827}  & \textbf{25.12}  & \textbf{0.548}  & \textbf{0.085} & 17.444 \\ \hline
0.01    & 0.821  & 24.73  & 0.579  & 0.088 & 17.735 \\ \hline
0.1     & 0.826  & 24.94  & 0.561  & 0.086 & \textbf{16.834} \\ \hline
0.5     & 0.821  & 24.78  & 0.600  & 0.091 & 17.233 \\ \hline
1.0     & 0.815  & 24.51  & 0.594  & 0.088 & 16.997 \\ 
\hline
\end{tabular}}
\label{table:a}
\end{table}

\begin{table*}[!t]
\centering
\captionsetup{font=footnotesize,labelsep=period}
\caption{Impact of individual components. We evaluate the performance gains of adding the auxiliary depth branch (Control 2) and Frequency Loss. Values in parentheses denote the improvement over the single-branch baseline. Results show that both modules contribute comparably and synergistically to the final image quality.
}
\renewcommand{\arraystretch}{1.2}
% \resizebox{\linewidth}{!}{
\begin{tabular}{ccc:ccccc}
\hline\hline
\multicolumn{3}{c:}{\textbf{Metric}} & \multirow{2}{*}{SSIM ($\uparrow$)}   & \multirow{2}{*}{PSNR (dB) ($\uparrow$)}   & \multirow{2}{*}{MSE (e-02) ($\downarrow$)} &  \multirow{2}{*}{LPIPS $\downarrow$} & \multirow{2}{*}{FID $\downarrow$}  \\
\cmidrule(lr){1-3}
Control 1 & Control 2 & Freq Loss & & & & & \\
\hline
\checkmark & &            
& 0.685           & 19.32       & 1.308         & 0.222         & 54.060  \\
\checkmark & \checkmark 
& & 0.795 (+0.110)& 23.42 (+4.1)& 0.709 (-0.599)& 0.104 (-0.118)& 18.573 (-35.487) \\
\checkmark & & \checkmark 
& 0.804 (+0.119)& 23.82 (+4.5)  & 0.653 (-0.655)& 0.100 (-0.122)& 19.558 (-34.502)\\
\checkmark & \checkmark & \checkmark
& 0.827 (+0.032)  & 25.12 (+1.3)& 0.548 (-0.105)& 0.085 (-0.015)& 17.444 (-1.129)\\
\hline\hline
\end{tabular}
% }
\label{tab:ablation_control}
\end{table*}
Building upon the optimized Dual-Control architecture ($w=0.3$), we further investigated the impact of the Frequency-Domain Supervision by varying the regularization weight $\alpha$. 
We explored a logarithmic scale $\alpha \in \{0.0001, 0.001, 0.01, 0.1, 0.5, 1.0\}$ to determine the optimal balance between spatial diffusion objectives and spectral consistency.
The results, presented in Table~\ref{table:a}, demonstrate the efficacy of the proposed frequency loss. 
The introduction of spectral supervision boosting the SSIM from $0.795$ (the best performance in the previous ablation without frequency loss) to a peak of 0.827. 
This confirms that explicit frequency-domain constraints effectively guide the model to recover high-frequency diffraction patterns that pixel-space objectives alone often neglect.
Furthermore, the model exhibits remarkable robustness to variations in $\alpha$. We observe that performance remains stable across orders of magnitude (from $\alpha=0.0001$ to $0.1$), with SSIM values consistently staying above $0.82$. 
This suggests that the frequency loss provides a consistent, complementary gradient direction that does not conflict with the spatial denoising objective. 
Although $\alpha=0.1$ achieves the lowest FID ($16.834$), we prioritise structural fidelity for scientific accuracy and select $\alpha=0.001$ as the final hyperparameter, as it delivers the highest structural similarity (SSIM: $0.827$) and peak signal-to-noise ratio (PSNR: $25.12$).

% \subsubsection{Influence of Sampling Steps}
\subsubsection{Impact of different modules}
% 这个实验验证了不同modules的作用，包括control1 and control2 对ControlNet的作用，以及Freqency loss的作用。Table VI里列出了加入每个module以后的结果以及相对于上一个模型的提升值。从我们的五个evaluation metrics里可以看出相对于controlNet，control2条件的加入和Frequency loss的加入对生成图片的质量的提升程度几乎相同，例如control2使得SSIM提升0.11，PSNR提升4.1，Frequency loss使得SSIM提升0.119，PSNR提升4.5。在这之后，control2和frequency loss的共同加入（对应最后一行的结果）又另外的提升了生成图片的质量。这个实验验证了每个module对我们Du-FreqNet的作用。

To isolate the contribution of each module, we performed an ablation study starting from the single-branch baseline (Control 1). Table~\ref{tab:ablation_control} reveals that adding the depth-specific Control 2 branch ($+0.110$ SSIM) and enforcing the Frequency Loss ($+0.119$ SSIM) yield comparable, significant performance boosts. This indicates that explicit geometric cues and spectral constraints are equally critical for resolving diffraction details. The full Du-FreqNet integrates both modules to achieve peak fidelity, demonstrating that these components are synergistic, one ensuring spatial structure, the other enforcing optical consistency.

\section{Conclusion}
In this paper, we presented Du-FreqNet, a physics-informed generative framework designed to address the data scarcity bottleneck in optical microrobotic perception. 
By synergising a dual-control architecture with an adaptive frequency-domain supervision mechanism, our model effectively decouples volumetric geometric priors from depth-dependent optical diffraction effects. 
This approach overcomes the inherent spectral bias of standard diffusion models, enabling the synthesis of high-fidelity microscopy images that are both structurally accurate and physically consistent. 
Extensive experiments demonstrate that Du-FreqNet not only outperforms state-of-the-art GAN baselines in synthesis quality but also generalizes effectively to unseen poses. Crucially, the performance in downstream Sim-to-Real tasks confirms the practical utility of our synthetic data for training robust robotic perception systems. 
Our findings suggest that integrating explicit optical physics into data-driven generative models offers a promising direction for automating intelligent microsystems.

\bibliographystyle{IEEEtran}
\bibliography{ref}

\end{document}